\pdfoutput=1
\documentclass[11pt]{article}

\usepackage[preprint]{acl}

\usepackage{times}
\usepackage{latexsym}

\usepackage{booktabs}
\usepackage{tcolorbox}

\usepackage{graphicx}
\usepackage{lipsum}
\usepackage{subcaption}
\usepackage{array}
\usepackage{svg}

\usepackage{amsmath}
\usepackage{amsfonts}
\usepackage{hyperref}

\usepackage{enumitem}
\setlist[itemize]{noitemsep, nolistsep}

\definecolor{correct}{RGB}{150, 200, 50}

\usepackage[T1]{fontenc}
\usepackage[utf8]{inputenc}

\usepackage{microtype}

\usepackage{inconsolata}

\usepackage{graphicx}

\title{A Tale of Trust and Accuracy: Base vs. Instruct LLMs in RAG Systems}

\author{
 \textbf{Florin Cuconasu\textsuperscript{1}\thanks{Correspondence to \href{mailto:email@domain}{cuconasu@diag.uniroma1.it}}},
 \textbf{Giovanni Trappolini\textsuperscript{1}},
 \textbf{Nicola Tonellotto\textsuperscript{2}},
 \textbf{Fabrizio Silvestri\textsuperscript{1}}
\\
 \textsuperscript{1}Sapienza University of Rome,
 \textsuperscript{2}University of Pisa
}

\begin{document}
\maketitle
\begin{abstract}
Retrieval Augmented Generation (RAG) represents a significant advancement in artificial intelligence combining a retrieval phase with a generative phase, with the latter typically being powered by large language models (LLMs). The current common practices in RAG involve using ``instructed'' LLMs, which are fine-tuned with supervised training to enhance their ability to follow instructions and are aligned with human preferences using state-of-the-art techniques.
Contrary to popular belief, our study demonstrates that base models outperform their instructed counterparts in RAG tasks by 20\% on average under our experimental settings. 
This finding challenges the prevailing assumptions about the superiority of instructed LLMs in RAG applications. Further investigations reveal a more nuanced situation, questioning fundamental aspects of RAG and suggesting the need for broader discussions on the topic; or, as Fromm would have it, ``Seldom is a glance at the statistics enough to understand the meaning of the figures''. \footnote{Translated from \href{https://www.mimesisedizioni.it/libro/9791222303000}{I cosiddetti Sani.}} \footnote{The code and data will be made available at \href{https://github.com/florin-git/Base-vs-Instruct-LLMs-in-RAG-Systems}{github.com/florin-git/Base-vs-Instruct-LLMs-in-RAG-Systems}}
\end{abstract}

\section{Introduction}

\begin{figure*}[htbp]
    % \centering
    \begin{tabular}{|p{0.47\textwidth}|p{0.47\textwidth}|}
        \hline
        \centering
        \textbf{Llama 2 7B} \arraybackslash  & \centering \textbf{Llama 2 7B-Chat + Template} \arraybackslash \\
        \hline
        {\emph{} \newline
        You are given a question and you MUST respond by EXTRACTING the answer from one of the provided documents. If none of the documents contain the answer, respond with NO-RES.} & \emph{[INST] <<SYS>>} \newline
        You are given a question and you MUST respond by EXTRACTING the answer from one of the provided documents. If none of the documents contain the answer, respond with NO-RES. \newline
        \emph{<</SYS>>} \\
        \hline
        \multicolumn{2}{|p{0.94\textwidth}|}{
            \textbf{Documents:}
            
            \textbf{Document [1]}(Title: Batman Returns) the Penguin. We didn't really officially cast it, but for a short nasty little guy, it's a short list. I ended up writing the character for Danny DeVito. \textcolor{correct}{Burgess Meredith} (who portrayed the Penguin in the 1960s TV series "Batman") was cast for a little cameo as Tucker Cobblepot...
            
            \textbf{Document [2]}(Title: Batman: Mystery of the Batwoman) This is the only time in the DC animated universe that Paul Williams did not voice the Penguin...
            
            \textbf{Document [3]}(Title: The Penguin's a Jinx) The Penguin goes to Wayne Manor and returns the actress. He then uses his gas-umbrella to knock out anyone inside the statues...\newline

            \textbf{Question:} Who played the part of `The Penguin' in the TV series `Batman'?
        } \\
        \hline
        \textbf{Answer:} \textcolor{correct}{Burgess Meredith} & 
        
        \textbf{Answer:} \emph{[/INST]} Based on the provided documents, the answer is \textcolor{red}{Danny DeVito} \\
        \hline
    \end{tabular}
    \caption{\textbf{Base vs. Instruct + Template under Task Instruction I on TriviaQA}. 
    The figure presents a comparison between the responses generated by two versions of the Llama 2 7B model: the base version and the instruct + template version. Each version is tasked with answering the same question based on the provided documents. The base model correctly identifies the answer as ``Burgess Meredith'', whereas the instruct + template version incorrectly attributes the answer to ``Danny DeVito''. \emph{Italic} text denotes the template.}
    \label{fig:base_vs_instruct}
\end{figure*}

Retrieval Augmented Generation (RAG)~\cite{lewis2020retrieval} is an innovative approach that enhances the capabilities of Large Language Models (LLMs) by integrating retrieval mechanisms into the generative process. At its core, RAG operates by retrieving relevant information from a vast corpus of data and then generating coherent and contextually enriched responses based on this retrieved information. This dual process not only improves the accuracy and relevance of the generated content but also addresses some of the inherent limitations of standalone generative models, such as hallucinations~\cite{huang2023survey} and context drift~\cite{wang2022measure}.
The significance of RAG in natural language processing and artificial intelligence cannot be overstated. As the demand for more sophisticated and context-aware AI systems grows, the ability to generate information that is both accurate and contextually relevant becomes crucial~\cite{gao2024retrievalaugmented}. 
RAG achieves this by leveraging the vast amount of information available, ensuring that the outputs of the models are informed by up-to-date and contextually appropriate data. This has profound implications for various applications, including conversational AI, information retrieval, and automated content generation~\cite{shuster2021retrieval,wang2024unims}.
Furthermore, RAG represents a paradigm shift in how we think about and utilize LLMs. Instead of relying solely on the generative power of these models, RAG harnesses the complementary strengths of retrieval systems. This synergy enables the creation of AI systems that are not only more knowledgeable but also more reliable and versatile in their applications \cite{izacard2021leveraging,zhu2024retrievalaugmented}.

%%%%%%%%%%%%%%%%%%%%%%%%%%%

LLMs are the key component in RAG systems. They are initially pre-trained on the task of next token prediction~\cite{radford2018improving}, where the LLM learns to predict the next word in a sequence based on the context provided by the preceding words. This extensive pre-training phase involves processing vast amounts of text data, enabling the model to acquire a broad understanding of language, syntax, semantics, and general knowledge.
We call this the ``base'' version.
Following this pre-training phase, LLMs typically undergo two stages of refinement to enhance their performance and usability, whose output we call the ``instruct''. 
The first stage is that of supervised instruction fine-tuning (SFT), where the goal is to teach the model to follow instructions passed through the prompt.
This is carried out by supervised fine-tuning, where the model is trained on a curated dataset incorporating instructions, with the specific goal of improving its ability to follow the specified instructions~\cite{taori2023alpaca}.
% The first stage focuses on teaching the LLM to generate responses or perform specific tasks according to some instructions included in their prompt. This is carried out by supervised fine-tuning, where the model is trained on a curated dataset incorporating instructions, with the specific goal of improving its ability to follow the specified instructions.
The second stage is alignment with human preferences, often referred to as reinforcement learning from human feedback (RLHF)~\cite{ouyang2022training} or similar methods~\cite{rafailov2023direct,hong2024orpo,rahman2022robust}. During this phase, the model's outputs are adjusted to better align with human preferences and values. This usually involves iterative processes where human evaluators provide feedback on the model's responses, and the model is further fine-tuned to produce outputs that are not only accurate but also contextually appropriate and aligned with human expectations.
%%%%%%%%%%%%%%%
In practice, the ``instruct'' versions of LLMs are the go-to models widely used for RAG tasks~\cite{liu2023lost,langchain2023,dspy2023}. 
Moreover, these instruct models often come with a ``template'', specific usage guidelines that provide a structured approach to utilizing the system effectively. 
% These templates should serve as a framework to ensure consistent and optimal performance across different applications.

In this paper, we conduct a principled evaluation of instruct models and their accompanying templates against their base versions in a RAG setting.
%—those models that have only undergone the initial pretraining phase.
% Our methodology involves rigorous testing to compare the performance of both model types in the context of RAG.
Surprisingly, our results reveal that base models, without the additional instruct-specific fine-tuning, outperform the instruct models on the task of RAG in our experimental setting. This finding challenges the prevailing assumption that instruct models are inherently superior for such tasks.
Upon further investigation, we uncover that the situation is more nuanced, 
% than initially apparent, 
with various factors contributing to the observed performance differences. 
Our study stimulates a broader discussion on RAG, its methodologies, and evaluation procedures employed to advance the state of the art in this field. 

In summary, our contributions are:

\begin{itemize}
    \item Performance Evaluation: We conduct a principled evaluation comparing instruct models and their templates against base models in the context of RAG, revealing that base models outperform instruct models.
    \item Nuanced Insights: Through detailed analysis, we uncover the complexities and nuances that influence the performance of RAG systems.
    \item Pathway for Future Research: Our findings challenge existing assumptions and stimulate further discussion on RAG's state of the art, helping the development of more effective and reliable systems.
\end{itemize}

\section{Background}

% In this section, we will examine the steps required to train a large language model, including pre-training, instruction fine-tuning, and alignment; followed by the foundations of RAG.

In this section, we will explore the steps involved in training a large language model, including pre-training, instruction fine-tuning, and alignment. We will then discuss the fundamentals of RAG.

\subsection{LLM Training}

Here, we illustrate the training processes of LLMs, which consist of at least three main steps: pre-training, which is done on the task of next token prediction, instruction fine-tuning, and preference alignment. 

\paragraph{Pre-training.}

Pre-training for large language models \cite{radford2018improving} involves an extensive unsupervised learning phase, where the model is exposed to a large corpus of text data to learn the underlying statistical properties of natural language. This process employs the next token prediction task, where the model is conditioned on a sequence of tokens $w_{1:i-1}$ and trained to predict the subsequent token $w_i$, i.e., $p(y) = \prod_i^n p_\theta(w_i | w_{1:i-1})$.
By iteratively processing vast and diverse textual datasets, the model learns linguistic patterns, including syntactic structures, semantic relationships, and contextual dependencies. 
% This phase leverages large-scale datasets such as books, articles, and web content, facilitating the development of a robust and comprehensive language representation. 
This process delivers what is commonly called a ``base'' model. This first model is usually improved with further specialized training, as we will see now.

\paragraph{Instruction Fine-Tuning.}

Instruction fine-tuning is the part of the training process of large language models aimed at enhancing their ability to follow specific directives and perform specialized tasks \cite{taori2023alpaca}. 
This phase involves further training of the base model on curated datasets that include explicit instructions paired with corresponding responses. 
The objective is to teach the model how to interpret and execute various types of commands or queries effectively.
During instruction fine-tuning, the model is exposed to a wide range of examples that illustrate how to respond to different types of prompts, from answering questions to generating summaries and more complex task-oriented interactions. 
% This training process helps the model develop a better understanding of how to follow human-like instructions and generate contextually appropriate and accurate responses.
The primary purpose of instruction fine-tuning is to improve the model's usability and performance in real-world applications. 
By aligning the model's behavior with the specific needs and expectations of users, instruction fine-tuning ensures that the model can handle diverse and nuanced tasks more effectively, enhancing its overall functionality and practical utility.
This process might involve using an instruction template, where specific patterns and structures for providing instructions and expected responses are standardized to facilitate consistent and effective learning.
It is worth noticing that this is still achieved with traditional supervised learning, unlike the next phase we are considering.

\paragraph{Aligning to Human Preferences.}

The final step, which might include multiple sub-steps, is aligning LLMs to human preferences.
Aligning AI systems ensures that these systems can effectively and ethically interact with humans, providing outputs that are useful, respectful, and culturally sensitive.
Human preferences are not easily integrated into a differentiable loss function, necessitating specialized techniques. The most popular are Direct Preference Optimization (DPO) \cite{xu2024dpo} and Reinforcement Learning from Human Feedback (RLHF) \cite{ouyang2022training}.
These techniques solve the following optimization problem:
\begin{align*}
J_r(\pi_\theta) = \mathbb{E}_{\mathbf{x}, \mathbf{y}} \left[ r(\mathbf{x}, \mathbf{y}) - \beta \log \frac{p_\theta(\mathbf{y} \mid \mathbf{x})}{p_{ref}(\mathbf{y} \mid \mathbf{x})} \right]
\end{align*}
The LLM $p_\theta(\mathbf{y} \mid \mathbf{x})$ is here considered as a policy that following an instruction $x$ generates a text response $y$.
The reward function $r(\mathbf{x}, \mathbf{y})$ reflects human preferences. It takes a prompt $\mathbf{x}$ and the corresponding response $\mathbf{y}$ as input and outputs a scalar value.
The reference model $p_{ref}$ is used to regularize the LLM model $p_\theta$ with Kullback–Leibler divergence. The hyper-parameter $\beta$ is a constant to control the degree of regularization.
RLHF solves this problem in a two-step procedure; first, it learns a reward model, and then it uses this learned reward model to optimize the equation above.
DPO instead skips the first step and directly solves a less general problem by optimizing the policy $p_\theta$ over preference data.
Despite the importance of aligning large language models with human preferences, the inner workings of these alignment techniques remain largely unclear. 
Recently, \cite{hong2024orpo} showed that applying RLHF can decrease the expected reward of a model but still improve its performance.
Currently, it remains unclear which technique is superior, with experimental evidence showing that it depends on the particular application (like DPO being bad at coding \cite{xu2024dpo}).
There is a significant effort in the scientific community to understand and improve these methods, aiming to develop more transparent and reliable alignment processes. 
In the remainder of this paper, we will try to measure its impact on RAG.

\subsection{RAG}

RAG \cite{lewis2020retrieval} is a hybrid approach that integrates information retrieval mechanisms with generative language models to enhance the quality and relevance of generated content. In RAG, a retrieval component first identifies and retrieves relevant documents or data from a large corpus based on a given query or context. This retrieved information is then used as supplementary input for the generative model, which synthesizes a response or generates new text informed by the retrieved data.
Specifically, given a corpus of documents $D$ and a query $q$, a retriever is employed to retrieve the set of documents $D_k$ that are more relevant to the query. In the dense setting, like the one we employ, this is achieved through a neural bi-encoder that independently encodes queries and documents in a learned latent vector space where a similarity score, like the cosine similarity, is used to extract the most similar documents, formally: $sim(q, d_i) \propto \overrightarrow{q} \cdot \overrightarrow{d_i}$,
% \begin{align*}
% sim(q, d_i) \propto \overrightarrow{q} \cdot \overrightarrow{d_i},
% \end{align*}
where $d_i$ is the i-th document, while $\overrightarrow{q}$ and $\overrightarrow{d_i}$ are the embedding of the query and i-th document, respectively.
The top-$k$ documents are then passed to a generative model, like an LLM.
The LLM takes an input, once again, the query $q$ and the set of most relevant documents $D_k$. The problem is then formulated as generating the answer to the query, conditioning on the query, and marginalizing over the retrieved documents:
\begin{align*}
    P_{\text{RAG}} (y|q) \approx \prod_{i}^{N} \sum_{d \in \mathcal{D}_k} p_{\eta} (d | q)  p_{\theta} (y_i | q, d, y_{1:i-1}),
\end{align*}
where $p_{\eta}(d \mid q)$ is the retrieval component that provides a (truncated) probability distribution for the top-scoring documents, and $p_{\theta}(y_i \mid q, d, y_{1:i-1})$ is a probability distribution parameterized by $\theta$ that generates the current token based on the previously generated tokens, the query, and the retrieved document.

\section{Experimental Methodology}

In this paper, we aim to achieve multiple objectives.
Firstly, we seek to determine whether base models outperform their instruction versions in the context of RAG.
Secondly, we investigate the underlying factors affecting RAG models' performance and the impact of the additional training techniques (i.e., SFT and Alignment) on these systems.
To this end, we set up a series of rigorous experiments to methodically evaluate and compare the performance and behavior of base models versus their instruct-tuned counterparts, providing a comprehensive understanding of their respective advantages and limitations in RAG tasks.

\subsection{Task Instructions}

We perform our experimenting with two task instructions.
Task instruction I (Figure \ref{fig:base_vs_instruct}) requires the LLM to ``extract'' the answer from one of the provided documents, reflecting the extractive nature of the QA datasets.
Additionally, the models are tasked to respond with \emph{NO-RES} if the answer is not present in the retrieved documents, testing their \emph{negative rejection} capabilities \cite{chen2024negativerejection}.
To check if the model is really using the provided context and it is aware of it, we also employ another task instruction.
% Task instruction II extends the first instruction by requiring the model to provide evidence of its response from the given context with a \emph{Proof}. 
Task instruction II builds on the first instruction by asking the model to provide evidence from the given context to support its response with a \emph{Proof}.
This instruction includes a one-shot example to demonstrate the expected format of the model's answer, as visualized in Figure \ref{fig:base_vs_instruct_proof}.

% In a RAG setting, the number of possible task instructions and prompts that can be adopted is potentially infinite. In this case, we consider two types of task instructions that can be generally employed. Task instruction I (Figure \ref{fig:base_vs_instruct}) requires the LLM to ``extract'' the answer from one of the provided documents, reflecting the extractive nature of the QA datasets. Moreover, the models are tasked to respond with \emph{NO-RES} if the answer is not present in the retrieved documents, testing their \emph{negative rejection} capabilities \cite{chen2024negativerejection}.

% Task instruction II extends the first instruction by requiring the model to provide evidence of its response from the given context with a \emph{Proof}. This instruction includes a one-shot example to demonstrate the expected format of the model's answer (refer to Figure \ref{fig:base_vs_instruct_proof}).

The full prompt is composed of the task instruction, to which the retrieved documents and the query are added.
% To the task instruction, it is then concatenated the retrieved documents and the query. 
Documents are ordered by ascending similarity score to position the high-similarity documents nearest to the query, according to the insights of \citet{liu2023lost}.

\subsection{Instruct Templates}

When fine-tuning LLMs to create their instruct versions, specific prompt templates are used during training. 
These templates are designed to clearly distinguish between model responses or task instructions and user inputs.
This distinction is marked through the use of special tokens. 
For instance, Llama 3 utilizes \emph{<|begin\_of\_text|>}, and Mistral uses \emph{[INST]} to specify the beginning of instructions.
Despite the use of these templates for instruct LLMs, their effects on model performance, when removed from the typical conversational setting and applied to a rag setting, remain understudied. 
% Specifically, it is unclear how these instruct models perform in generic question-answering tasks where no assistant-user dialogue format is required. 
Our study explores this gap by evaluating the performance of instruct models with and without their standard chat templates. % in a non-dialogue QA setting.

\subsection{Datasets and Models}
In our experiments, we use two open-domain question-answering datasets: the open version of Natural Questions (NQ-open) \cite{kwiatkowski2019natural, nq-open} and TriviaQA-unfiltered \cite{triviaqa}. 
% These two tasks are those who are mostly used in RAG experiments (CITAZIONE), we defer to a future work experiments with other tasks. 
For each query, the Contriever \cite{izacard2021contriever} is used to retrieve the most similar documents from the English Wikipedia corpus. 
The retriever's performance is discussed in Section \ref{sec:retriever_appendix}.
For generation, we employ several LLMs in both their base and instruct/chat versions: 
\textbf{(a)} Llama 2 7B and Llama 2 7B-Chat;
\textbf{(b)} Llama 3 8B and Llama 3 8B-Instruct;
\textbf{(c)} Falcon 7B and Falcon 7B-Instruct;
\textbf{(d)} Mistral 7B and Mistral 7B-Instruct;
% \begin{itemize}
%     \item Llama 2 7B and Llama 2 7B-Chat
%     \item Llama 3 8B and Llama 3 8B-Instruct
%     \item Falcon 7B and Falcon 7B-Instruct
%     \item Mistral 7B and Mistral 7B-Instruct
% \end{itemize}
Models are quantized at 4-bit.
% in order to allow us to run more experiments with the limited amount of compute resources we had available.
For clarity, we will refer to the Llama 2 chat version collectively as ``instruct''.
All models utilize greedy decoding, and the maximum response length is tailored to the requirements of each dataset. 
Under Task Instruction I, the response limit is set to 15 tokens for the NQ dataset, which demands short responses of no more than 5 tokens, and up to 50 tokens for the TriviaQA dataset to accommodate potentially longer answers. 
For Task Instruction II, which requires the proof, the maximum response length is increased to 200 tokens.

\subsection{Evaluation}

\paragraph{Accuracy.}

Accuracy is the main metric adopted to evaluate the models' responses. In particular, it is checked whether one of the ground truth answers of the dataset is contained in the generated response after applying a normalization process. This normalization involves the removal of punctuation and articles to ensure that the answer is not unfairly penalized by minor discrepancies in formatting. 
This type of evaluation may have flaws since it can incorrectly mark a correct response as incorrect if the ground truth answer is not fully contained in the generated response, even after normalization. Nevertheless, given that the employed datasets usually require short answers (e.g., answers in NQ-open are at most 5 tokens long) that can be extracted from the provided documents, adopting accuracy may be generally considered a suitable metric.

% As stated by \citet{cuconasu2024power}, this type of evaluation may have flaws, since it can incorrectly mark a correct response as incorrect if the ground truth answer is not fully contained in the generated response, even after normalization. Nevertheless, given that the employed datasets usually require short answers (e.g., answers in NQ-open are at most five tokens long) that can be extracted from the provided documents, adopting accuracy may be generally considered a suitable metric.

% Moreover, TriviaQA presents many different potential answers for each query, reaching in some cases up to 400 possible answers. This diversity helps reduce the number of times a generated response is considered incorrect even though it is semantically correct. This is evident in the results (Section \ref{sec:results}), where the accuracy on TriviaQA is greater in absolute terms compared to the NQ scores.

\paragraph{Negative Rejection.}

As shown in Figure \ref{fig:base_vs_instruct}, LLMs are tasked with responding with \emph{NO-RES} when the provided documents do not contain the necessary knowledge to answer the query. This approach helps assess the models' abilities to understand the task instructions and correctly refuse to answer when the information is not present, thereby reducing the occurrence of hallucination \cite{zhang2023sirens}.
The \emph{negative rejection} ability can be measured with the \emph{rejection rate}, which was introduced by \citet{chen2024negativerejection}. It is computed as the number of times the model answers with \emph{NO-RES} when the documents indeed lack the required information, divided by the total number of such instances.
High rejection rates indicate that the model effectively avoids generating potentially incorrect or misleading answers.
%, which is essential for maintaining the trustworthiness of AI systems. 

% \florin{Da vedere se mettiamo questa parte}
% In addition to negative rejection, we also propose to consider \textbf{false negative rejection}, which assesses the model's ability to effectively use the information in the documents. The associated \emph{false negative rejection rate} measures instances where the model incorrectly responds with ``NO-RES'' despite the answer being present in the documents. Balancing false negative rejection and negative rejection rates is critical to optimize the model's performance, ensuring it neither over-rejects valid answers nor under-rejects when the information is missing.

\section{Results}
\label{sec:results}

In this section, we present the results for three types of models—base, instruct, and instruct + template—evaluated under various task instructions across different datasets, as detailed in the previous section.

\begin{table*}[ht]
\centering
\caption{\textbf{Task Instruction I Accuracy} on 
NQ and TriviaQA. The abbreviations \emph{C} and \emph{I} denote the \emph{Chat} and \emph{Instruct} versions of the instruct models, respectively. The suffix \emph{T} indicates instruct models using a \emph{Template} to structure their responses.
Accuracies are reporting at different levels of retrieved documents.
With the, partial, exception Mistral, all base models outperform their instruct counterparts by a considerable margin.}
\label{tab:nq_trivia_first}
\resizebox{2.09\columnwidth}{!}{
\begin{tabular}{@{}l|ccccccc|rrrrrrr@{}}
\toprule
 &
  \multicolumn{7}{c|}{\textbf{\begin{tabular}[c]{@{}c@{}}NQ\\ \# Retrieved Documents\end{tabular}}} &
  \multicolumn{7}{c}{\textbf{\begin{tabular}[c]{@{}c@{}}TriviaQA\\ \# Retrieved Documents\end{tabular}}} \\ \midrule
\multicolumn{1}{c|}{\textbf{Model}} &
  1 &
  2 &
  3 &
  4 &
  5 &
  8 &
  10 &
  \multicolumn{1}{c}{1} &
  \multicolumn{1}{c}{2} &
  \multicolumn{1}{c}{3} &
  \multicolumn{1}{c}{4} &
  \multicolumn{1}{c}{5} &
  \multicolumn{1}{c}{8} &
  \multicolumn{1}{c}{10} \\ \midrule
Llama 2 7B &
  \textbf{23.88} &
  \textbf{24.71} &
  \textbf{27.83} &
  \textbf{29.53} &
  \textbf{30.22} &
  \textbf{31.01} &
  \textbf{31.46} &
  \textbf{55.85} &
  \textbf{57.15} &
  \textbf{59.28} &
  \textbf{60.40} &
  \textbf{61.24} &
  \textbf{62.93} &
  \textbf{63.89} \\
Llama 2 7B-C &
  16.06 &
  18.48 &
  18.62 &
  18.59 &
  19.21 &
  21.98 &
  21.08 &
  32.59 &
  34.63 &
  41.09 &
  42.95 &
  46.14 &
  49.19 &
  48.96 \\
Llama 2 7B-C + T &
  3.36 &
  1.21 &
  0.69 &
  0.48 &
  0.45 &
  0.73 &
  1.52 &
  23.35 &
  21.63 &
  21.13 &
  17.70 &
  18.05 &
  22.30 &
  25.68 \\ \midrule
Llama 3 8B &
  \textbf{27.03} &
  \textbf{30.22} &
  \textbf{30.53} &
  \textbf{31.08} &
  \textbf{29.08} &
  \textbf{29.49} &
  \textbf{28.70} &
  \textbf{44.64} &
  \textbf{53.48} &
  \textbf{56.78} &
  \textbf{59.59} &
  \textbf{58.97} &
  \textbf{64.93} &
  \textbf{65.05} \\
Llama 3 8B-I &
  8.52 &
  10.25 &
  15.40 &
  15.85 &
  22.57 &
  28.80 &
  28.25 &
  4.13 &
  2.62 &
  3.47 &
  4.44 &
  15.73 &
  51.90 &
  58.61 \\
Llama 3 8B-I + T &
  14.40 &
  17.34 &
  19.04 &
  15.30 &
  18.10 &
  20.35 &
  19.35 &
  16.45 &
  27.94 &
  31.32 &
  30.80 &
  34.18 &
  44.04 &
  50.12 \\ \midrule
Mistral 7B &
  \textbf{24.26} &
  \textbf{25.30} &
  25.72 &
  26.17 &
  27.66 &
  26.65 &
  28.87 &
  \textbf{58.67} &
  \textbf{59.15} &
  \textbf{58.87} &
  \textbf{60.49} &
  \textbf{62.02} &
  \textbf{64.16} &
  \textbf{65.92} \\
Mistral 7B-I &
  20.04 &
  24.99 &
  \textbf{26.69} &
  \textbf{30.56} &
  \textbf{31.67} &
  \textbf{33.33} &
  \textbf{34.82} &
  48.85 &
  52.31 &
  54.72 &
  55.90 &
  56.97 &
  59.06 &
  60.60 \\
Mistral 7B-I + T &
  18.17 &
  23.54 &
  19.14 &
  27.90 &
  27.45 &
  27.07 &
  26.79 &
  45.88 &
  50.22 &
  52.42 &
  53.73 &
  54.37 &
  57.56 &
  58.31 \\ \midrule
Falcon 7B &
  \textbf{17.13} &
  \textbf{18.97} &
  \textbf{21.15} &
  \textbf{21.08} &
  \textbf{21.95} &
  \textbf{22.64} &
  \textbf{22.33} &
  \textbf{41.64} &
  \textbf{43.11} &
  \textbf{43.69} &
  \textbf{44.22} &
  \textbf{45.60} &
  \textbf{46.35} &
  \textbf{48.24} \\
Falcon 7B-I &
  15.68 &
  17.72 &
  17.96 &
  19.21 &
  20.08 &
  20.04 &
  20.98 &
  33.19 &
  36.52 &
  36.46 &
  37.71 &
  38.42 &
  38.83 &
  39.33 \\ \bottomrule
\end{tabular}
}
\end{table*}

% \usepackage{booktabs}
\begin{table*}[ht]
\centering
\caption{\textbf{Task Instruction II Accuracy} on 
NQ and TriviaQA, where a \emph{Proof} is required. The abbreviations \emph{C} and \emph{I} denote the \emph{Chat} and \emph{Instruct} versions of the instruct models, respectively. The suffix \emph{T} indicates instruct models using a \emph{Template} to structure their responses.
Accuracies are reporting at different levels of retrieved documents.
In all cases considered, base models outperform their instruct counterparts by a considerable margin.}
\label{tab:nq_trivia_proof}
\resizebox{2.09\columnwidth}{!}{
\begin{tabular}{@{}l|ccccccc|ccccccc@{}}
\toprule
 &
  \multicolumn{7}{c|}{\textbf{\begin{tabular}[c]{@{}c@{}}NQ\\ \# Retrieved Documents\end{tabular}}} &
  \multicolumn{7}{c}{\textbf{\begin{tabular}[c]{@{}c@{}}TriviaQA\\ \# Retrieved Documents\end{tabular}}} \\ \midrule
\multicolumn{1}{c|}{\textbf{Model}} &
  1 &
  2 &
  3 &
  4 &
  5 &
  8 &
  10 &
  1 &
  2 &
  3 &
  4 &
  5 &
  8 &
  10 \\ \midrule
Llama 2 7B &
  \textbf{24.82} &
  \textbf{28.70} &
  \textbf{31.71} &
  \textbf{32.85} &
  \textbf{34.09} &
  \textbf{35.62} &
  \textbf{35.79} &
  \textbf{54.94} &
  \textbf{56.94} &
  \textbf{58.88} &
  \textbf{60.07} &
  \textbf{60.87} &
  \textbf{62.75} &
  \textbf{63.73} \\
Llama 2 7B-C &
  18.41 &
  24.96 &
  28.76 &
  30.22 &
  32.16 &
  33.16 &
  32.05 &
  41.86 &
  47.18 &
  49.53 &
  51.19 &
  52.59 &
  55.73 &
  56.00 \\
Llama 2 7B-C + T &
  1.59 &
  2.42 &
  3.88 &
  5.75 &
  8.45 &
  12.81 &
  12.91 &
  4.22 &
  6.55 &
  8.75 &
  10.15 &
  14.90 &
  20.69 &
  27.48 \\ \midrule
Llama 3 8B &
  \textbf{29.39} &
  \textbf{31.53} &
  \textbf{34.72} &
  \textbf{37.07} &
  \textbf{36.59} &
  \textbf{39.15} &
  \textbf{40.22} &
  \textbf{61.57} &
  \textbf{63.32} &
  \textbf{64.56} &
  \textbf{65.93} &
  \textbf{66.52} &
  \textbf{67.67} &
  \textbf{68.04} \\
Llama 3 8B-I &
  18.83 &
  23.57 &
  25.65 &
  27.97 &
  30.84 &
  34.13 &
  37.97 &
  42.67 &
  44.44 &
  47.11 &
  48.10 &
  48.71 &
  52.59 &
  56.48 \\
Llama 3 8B-I + T &
  11.70 &
  16.44 &
  6.40 &
  6.61 &
  11.60 &
  5.64 &
  0.69 &
  30.52 &
  36.89 &
  31.11 &
  33.17 &
  38.08 &
  33.69 &
  23.42 \\ \midrule
Mistral 7B &
  \textbf{24.82} &
  \textbf{30.74} &
  \textbf{33.75} &
  \textbf{35.17} &
  \textbf{36.83} &
  \textbf{40.15} &
  \textbf{40.15} &
  \textbf{57.98} &
  \textbf{60.49} &
  \textbf{62.18} &
  \textbf{63.70} &
  \textbf{65.23} &
  \textbf{67.04} &
  \textbf{67.44} \\
Mistral 7B-I &
  21.53 &
  27.66 &
  29.91 &
  32.85 &
  33.58 &
  36.45 &
  35.76 &
  48.51 &
  52.60 &
  54.79 &
  56.05 &
  56.91 &
  58.80 &
  59.54 \\
Mistral 7B-I + T &
  16.86 &
  18.10 &
  20.87 &
  22.43 &
  24.09 &
  24.44 &
  26.96 &
  33.73 &
  33.21 &
  35.47 &
  36.57 &
  34.98 &
  42.39 &
  48.45 \\ \midrule
Falcon 7B &
  \textbf{17.58} &
  \textbf{20.04} &
  \textbf{22.43} &
  \textbf{23.75} &
  \textbf{23.85} &
  \textbf{26.12} &
  \textbf{26.72} &
  \textbf{40.32} &
  \textbf{42.76} &
  \textbf{44.92} &
  \textbf{45.45} &
  \textbf{46.47} &
  \textbf{47.23} &
  \textbf{48.57} \\
Falcon 7B-I &
  16.10 &
  18.31 &
  18.93 &
  19.87 &
  20.91 &
  21.61 &
  21.30 &
  33.54 &
  35.69 &
  37.50 &
  38.09 &
  39.02 &
  41.08 &
  42.95 \\ \bottomrule
\end{tabular}
}
\end{table*}

\subsection{Evaluation on Task Instruction I}
In this initial set of experiments, we evaluate the models using Task Instruction I, pictured in Figure \ref{fig:base_vs_instruct}. 
Reported are the accuracies for each model/version combination at different levels of retrieved documents, that is, the number of documents added to the prompt of the LLM given the query.
Unexpectedly, we find that the base models always outperform their instruct counterparts (with one exception), as evidenced by results in Table \ref{tab:nq_trivia_first}.
Llama 2's base model outperforms its instruct counterpart (w/o template), averaged across retrieved documents, by 9.23 (48\%) and 17.88 (+42\%) points on NQ and Trivia QA, respectively.
Similarly, Falcon's base model is 1.94 (+10\%) and 7.48 (+20\%) points better.
Even more strongly, Llama 3's base model improves accuracy by 10.92 (+59\%) and 37.5 (+186\%); this is somewhat caused by Llama 3 reliance on its template; we'll examine this in Section \ref{sec:template}.
The only ``half'' exception is constituted by Mistral. 
In fact, Mistral's base model is -2.49 (-8\%) less accurate than the instruct version on NQ. However, it is still 5.83 (+10\%) more accurate on TriviaQA.
%

% The results for the NQ and TriviaQA datasets, presented in Table \ref{tab:nq_trivia_first}, show that the base models generally outperform the instruct versions. An exception is observed with the Mistral model on the NQ dataset, where the instruct version without a template is more effective, especially when the context includes more than three retrieved documents.

\begin{figure*}[htbp]
    \centering
    \includegraphics[width=\textwidth]{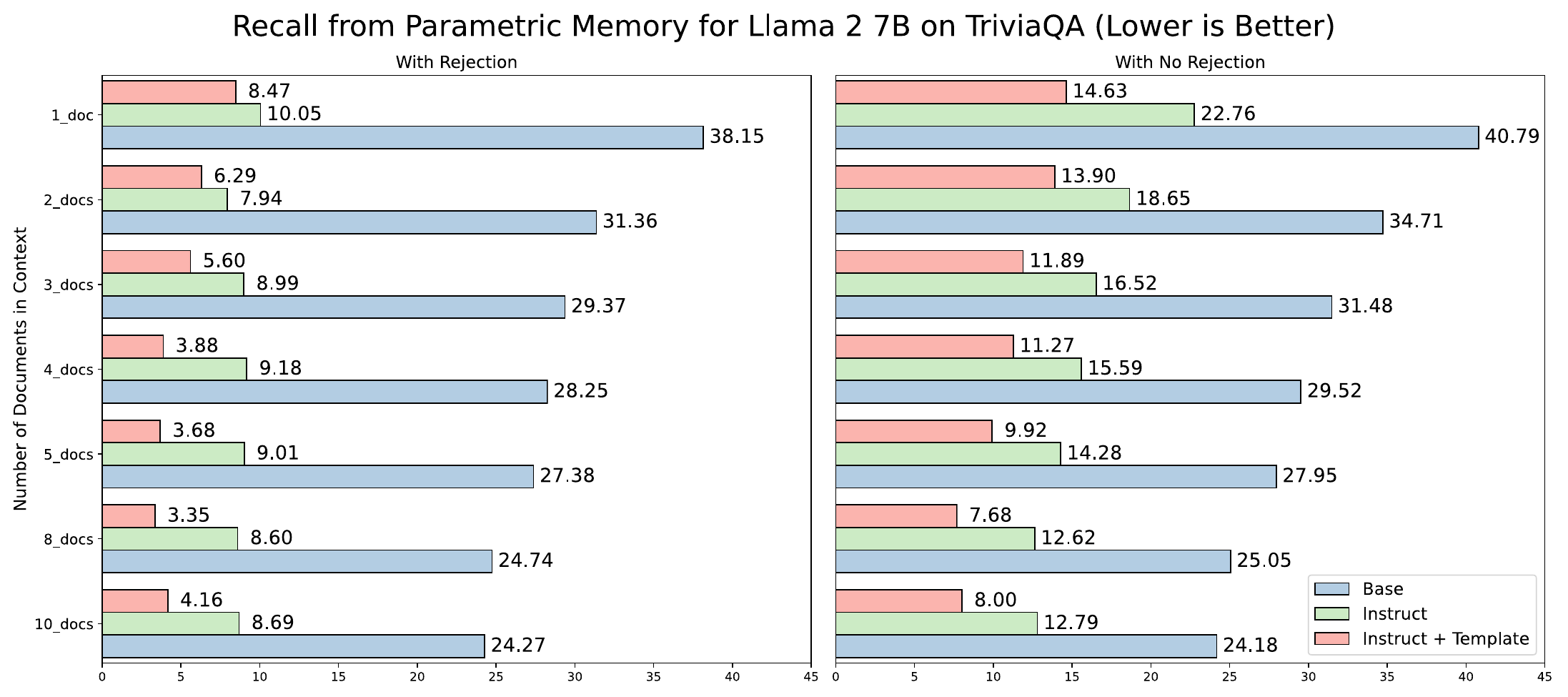}
  \caption{\textbf{Recalling from Parametric Memory - Llama 2 7B - TriviaQA}. 
  Reported is the recall from parametric memory rate, defined as the number of instances where the model correctly answers despite the retrieved documents not containing the correct answer, divided by the number of times the answer is not present in the context.
  \textbf{(left)} Task Instruction I as shown in Figure \ref{fig:base_vs_instruct}; \textbf{(right)} No Rejection setting, where we do not specify to answer with \emph{NO-RES} when the answer is not contained in the retrieved documents (example in Figure \ref{fig:base_vs_instruct_no_rej}). In this case, the parametric memory rate increases for both model versions.}
  % This figure shows the instances when the model responds correctly when the answer is not in context: on the left, under Task Instruction I as shown in Figure \ref{fig:base_vs_instruct}; on the right, when using Task Instruction I with No Rejection, as shown in Figure \ref{fig:base_vs_instruct_no_rej}.}
  \label{fig:llama2_canc}
\end{figure*}

\subsection{Evaluation on Task Instruction II}

Intrigued by the first set of results, we proceed to examine a new task instruction developed to test the model's abilities to ground their answers.
In this setting, models are required to provide a \emph{Proof}, a piece of evidence to substantiate their answers based on the information present in the context documents.
Examples of this setup are illustrated in Figure \ref{fig:base_vs_instruct_proof_mistral_nq} and \ref{fig:base_vs_instruct_proof} for NQ and TriviaQA, respectively.
Results can be seen in Table \ref{tab:nq_trivia_proof}.
We can immediately notice that there is a general upward shift in terms of accuracy for all models and settings; for instance, for Llama 2 base, it increases by 3.56 (+12\%) on average.  
This probably indicates that asking for the proof is in itself a form of prompt engineering.
Furthermore, we can observe that base models still outperform their counterparts.
Actually, these results are even stronger, as in this setting Mistral base achieves higher accuracy than the instruct versions by 3.41 (+10\%), too.

\begin{figure}[ht]
    \centering
    \includegraphics[width=\columnwidth]{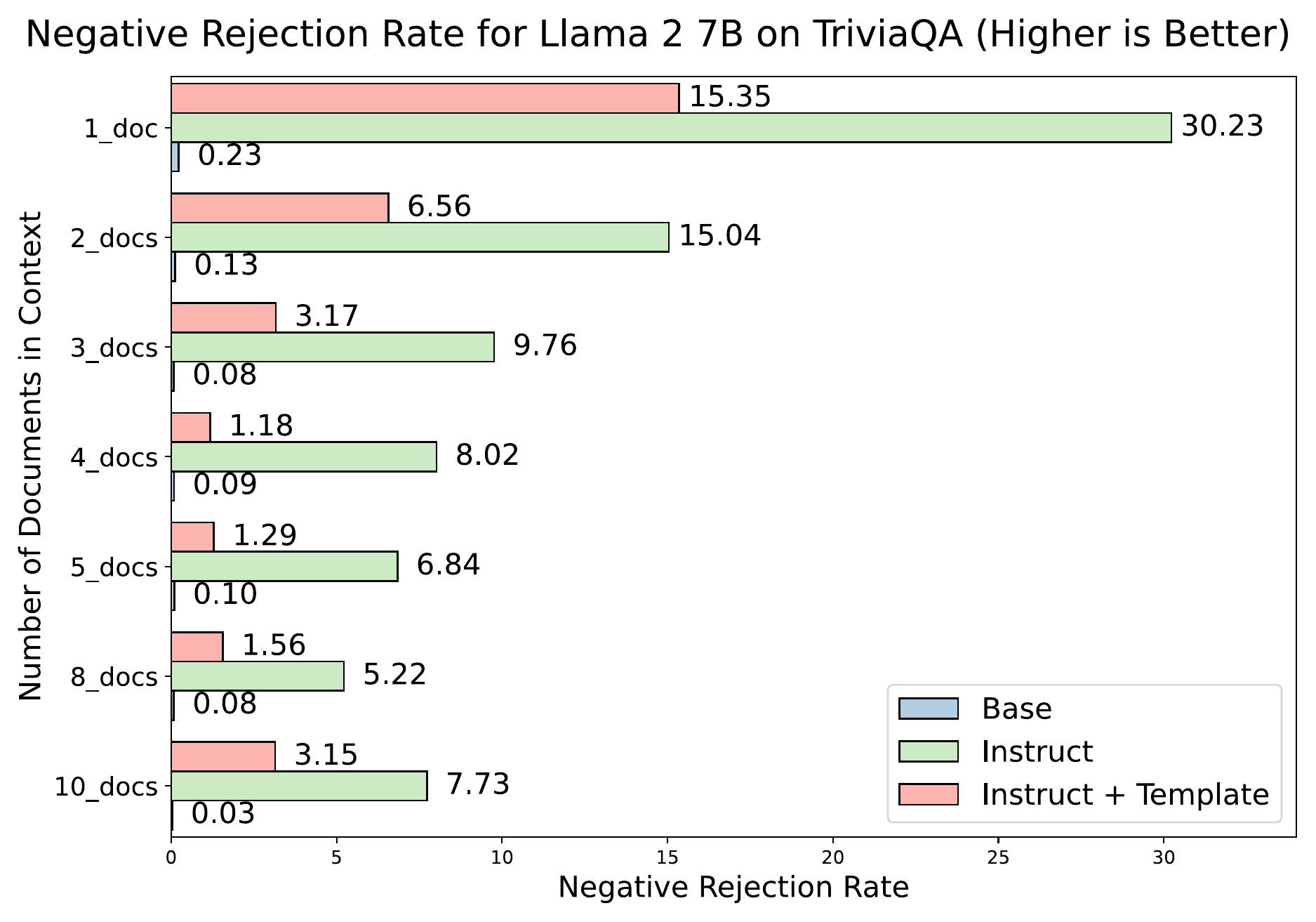}
  \caption{\textbf{Negative Rejection Rate - Llama 2 7B - TriviaQA.} Reported is the negative rejection rate, that is, the number of times the model answers \emph{NO-RES} when the correct answer is not in the context, divided by the number of times the answer is indeed missing.
  Instruct models are much more effective at detecting such cases and following the instructions provided.}
  \label{fig:llama2_neg_rej}
\end{figure}

\subsection{Instruct Models with Template}
\label{sec:template}
Our results highlight the difficulty that instruct models face in answering the question when the recommended template is used.
Investigating this issue, we find that even though the instructions demand short answers, models in this setting override this specification and produce overly verbose responses, damaging their accuracy.
% An intriguing finding from our experiments is the difficulty instruct models with templates face in extracting the correct answers from the context.
% The primary reason for their lower scores is verbosity, which leads to their answers being truncated due to the maximum generation length parameter.
% Despite the task instructions specifying brief or simple extraction responses, models with templates tend to generate longer answers.
This tendency may be linked to their fine-tuning and alignment for conversational purposes, where verbosity can be advantageous to assist users.
This results in template-less instruct models exhibiting higher accuracy than their templated counterparts.
% Interestingly, instruct models that do not employ the template used during the instruction fine-tuning phase 
% Specifically, the Llama 2 instruct model without a template strongly outperforms the corresponding with a template. 
For instance, in the NQ dataset using Task Instruction I, Llama 2's templated version barely achieves a 3\% accuracy rate, while it only surpasses 10\% when the context contains more than 8 documents under Task Instruction II.
An additional notable aspect is observed with Llama 3 instruct on Task Instruction I. When fewer than 4 documents are retrieved, the instruct model without a template fails to understand the prompt and generates random text. As the number of documents increases, however, its performance substantially improves, eventually surpassing the templated model.
This suggests that a larger input length might have a role in overruling learned behavior.

% This suggests that a larger input length aids in better understanding the prompts. Indeed, when using Task Instruction II, which consists of a longer prompt, due to the presence of the one-shot example, Llama 3 instruct maintains a score that is consistent with the base model.

% This problem is especially notable in the NQ dataset where the LLM’s generation stops at 15 tokens. However, task instructions for NQ do specify that answers should be concise, ideally extracting a maximum of 5 tokens from the context (Figure ??).

% In addition to this, instruct models with templates underperform compared to their non-template counterparts, also in the TriviaQA dataset, where the maximum token generation limit is 50 (Table \ref{tab:trivia_first}). Despite the task instructions specifying brief or simple extraction responses, models with templates tend to generate longer answers. This tendency may be linked to their fine-tuning and alignment for conversational purposes, where verbosity can be advantageous to assist users. However, when templates are removed, these models still manage to answer accurately, often outperforming their templated versions.

\section{Is Accuracy Sufficient?}

Section \ref{sec:results} clearly indicates that base models outperform instruct models on RAG.
But is that really the case? \emph{Are base models truly better than the instruct counterpart on RAG-like prompts?}
To answer this question, in this section, we go more in-depth in analyzing and comparing their behavior.
% In this section, we go more in-depth in analyzing and comparing base and instruct models. The central question is: are base models truly \emph{better} than the instruct counterpart on RAG-like prompts? 
First, we test the ability of these models to adhere to the task instructions.
In particular, whether they appropriately respond with \emph{NO-RES} when no relevant answer is present in the retrieved documents, which we call negative rejection rate.

% To determine what \emph{better} could mean, we first consider the conventional metric of accuracy. Results from previous experiments indicate that base models indeed perform superiorly in terms of pure accuracy scores. However, it remains unclear if these models thoroughly adhere to the task instructions, specifically whether they appropriately respond with \emph{NO-RES} when no relevant answer is present in the documents.
%, and if they respond correctly under such circumstances.

% To assess the availability of an answer within the provided documents, we compute the top-$k$ accuracy of the retriever. This metric evaluates how often the ground truth answer appears within the top-$k$ documents retrieved for a query. Scores can be seen in Table \ref{tab:contriever_acc}.

\subsection{Negative Rejection Rate}
\label{sec:neg_rej}

In Figure \ref{fig:llama2_neg_rej}, we plot the negative rejection rates for various configurations (base, instruct, and instruct + template) of Llama 2 7B. It is evident that in most cases, the models fail to comply with the instruction to answer with \emph{NO-RES} when the answer is absent. As an example, the instruct version of Llama 2 responds with \emph{NO-RES} only 30.23\% of the time when the answer is not in the one document context. The non-compliance to the task instruction is especially pronounced in the base model, which seldomly opts for rejection.
It is worth noticing that as the number of documents in the context increases, all models tend to respond less frequently with \emph{NO-RES}, suggesting both that a higher volume of documents might introduce more distracting documents \cite{shi2023large}, leading the LLM to respond but erroneously, and that a larger input length might overrule task instructions.

% Our observations reveal notable insights when examining instances where the ground truth answer is not present in the context documents. In Figure \ref{fig:llama2_neg_rej}, we plot the negative rejection rates for various configurations (base, instruct, and instruct + template) of Llama 2 7B. It is evident that in most cases, the models fail to comply with the instruction to answer with \emph{NO-RES} when the answer is absent. As an example, the instruct version of Llama 2 responds with \emph{NO-RES} only 30.23\% of the time when the answer is not in the one document context. The non-compliance to the task instruction is especially pronounced in the base model, which almost never opts for rejection. In contrast, the instruct versions tend to adhere more closely to the task instruction and attempt to reject answering. Additionally, as the number of documents in the context increases, all models tend to respond less frequently with \emph{NO-RES}, suggesting that a higher volume of documents might introduce more distracting documents \cite{shi2023large}, leading the LLM to respond but erroneously.

\subsection{Recall From Parametric Memory}

Next, we consider cases where the correct answer is not present in the provided documents, yet the model still responds accurately.
As illustrated in the left part of Figure \ref{fig:llama2_canc}, base models frequently manage to provide the correct answer even when there is none in the retrieved documents, suggesting that they ``know'' the answer from prior training.
We call this ``recall from parametric memory'' (by parametric memory, we mean knowledge learned during training and stored in the parameters of the model, as opposed to non-parametric memory provided in the context through retrieved documents).
Recall from parametric memory is not inherently problematic.
A user might choose to both fine-tune on proprietary data and use RAG to get the highest possible accuracy.
However, the specific instructions for this study emphasize that models should opt not to answer when the correct response is not evident in the documents.
Not following this guideline raises important questions about their reliance on internal knowledge versus contextual information, particularly in settings where accurate rejection of unanswerable questions is crucial.
% In this particular study, we aim to determine whether base models perform better simply because they do not adhere to these instructions, prompting us to continue our investigation.

% A particularly intriguing aspect of our analysis emerges when considering cases where the correct answer is not present in the provided documents, yet the model still responds accurately. As illustrated in the left part of Figure \ref{fig:llama2_canc}, base models frequently manage to provide the correct answer, suggesting that they ``know'' the answer from prior training.

% While possessing pre-existing knowledge is not inherently problematic — many queries might indeed be answerable through general knowledge — the specific instructions for this study emphasize that models should opt not to answer when the correct response is not evident in the documents. However, it appears that in a significant majority of these cases, base models are still able to recall the answer. This capability raises important questions about their reliance on internal knowledge versus contextual information, particularly in settings where accurate rejection of unanswerable questions is crucial.

\subsection{Evaluation with No Rejection}

Here we aim to determine whether base models perform better simply because they do not adhere to these instructions, prompting us to continue our investigation by removing the requirement to respond with \emph{NO-RES} in the prompt.
As shown in the right part of Figure \ref{fig:llama2_canc}, instruct models also demonstrate a capacity to recall from parametric memory, although less frequently than their base counterparts.
Moreover, evidence from Table \ref{tab:nq_trivia_no_rejection} (appendix) suggests a slight improvement in accuracy for instruct models under these modified conditions.
However, base models still outperform instruct ones.
The results indicate that the processes of supervised fine-tuning and alignment detrimentally impact the model's capabilities in RAG. 
Moreover, a tradeoff is observed between the trustworthiness of the model and its ability to perform RAG effectively. 
As alignment and fine-tuning efforts enhance the reliability and adherence to desired behaviors, they simultaneously constrain the model's flexibility and efficiency in RAG tasks, highlighting a critical balance between trustworthiness and functional performance.

% Previous analyses reveal that base models exhibit greater accuracy mainly due to their ability to ``remember'' the correct answers. Moreover, these models often fail to adhere to instructions to reject answering when the correct answer is not present in the context.

% Interestingly, as shown in the left part of Figure \ref{fig:llama2_canc}, instruct models also demonstrate a capacity to ``remember'' the correct answers, although less frequently than their base counterparts. This observation raises a pertinent question: do instruct models enhance their ``recalling'' capabilities when the requirement to respond with \emph{NO-RES} is removed?

% Evidence from Table \ref{tab:nq_trivia_no_rejection} suggests a slight improvement in accuracy for instruct models under these modified conditions compared to their performance shown in Table \ref{tab:nq_trivia_first}. However, despite these gains, instruct models generally do not outperform the base models, except in cases like Llama 3 and Mistral on the NQ dataset when more documents are provided.

% When examining instances where instruct models correctly answer questions without the presence of the answer in the documents, in the case of \emph{NO-RES} in the prompt, we can notice a modest improvement from the previous task instructions, as depicted in the right part of Figure \ref{fig:llama2_canc}.
% Overall, it appears that supervised fine-tuning and alignment reduce the instruct models’ propensity to rely on ``recalling'' capabilities, an ability that persists in base models.

\section{Related Works}

Recent studies have highlighted challenges and potential improvements in language models' use of non-parametric versus parametric memory in question-answering tasks. 
Several papers \cite{krishna2021hurdles, shi2024detecting, carlini2019, kandpal2023struggle, mallen2023trust} demonstrate that LMs often rely on memorized answers, capable of responding correctly even when presented with irrelevant documents. 
Similarly, other studies \cite{longpre2021entity, xie2024adaptive} observe that LMs continue to leverage their parametric knowledge despite prompt modifications with contrasting entities.
\citet{wu2024clasheval} describes this phenomenon as a balance between the model's inherent knowledge and its adherence to newly retrieved information, underscoring the ongoing challenge of enhancing model responsiveness to dynamic inputs. 
On enhancing reliance on provided content, \citet{zhang2024raft} introduced a training strategy that emphasizes evidence-based responses, similar to the \emph{Proof} mechanism in our QA tasks. 
This method has shown potential in improving model effectiveness by grounding responses in factual evidence, even though hallucination issues still remain an open problem \cite{zuccon2023chatgpt, gao2023enabling}.
\citet{cuconasu2024power} found instruct models to be slightly more effective, but theirs was a controlled setting in which the ground truth was always provided.
% Conversely, \citet{cuconasu2024power} found that adding random documents to prompts can surprisingly increase accuracy, particularly when the correct information is included. 
% They noted that in such controlled settings, instruct models, which are designed to utilize contextual information more effectively, outperform base models.
These insights collectively underline the intricate balance between leveraging learned knowledge and external data in improving QA systems, suggesting directions for future research in training strategies and model design.

% In another work, \citet{longpre2021entity} noticed that LMs overlay on their parametric knowledge despite modifying the prompt with contrastive entities.

% \citet{krishna2021hurdles} already encountered the problem of models that memorized answers in long-form QA. They showed that even replacing the retrieved documents with random ones the model is still able to answer correctly. 

% Instead, \cite{cuconasu2024power} studied a phenomenon where, by adding random documents in the prompt, the LLM seems to improve in accuracy. They also claimed that in the controlled setting where the gold document is available instruct models show superior performance than the base counterpart. 

% \cite{shi2024detecting} demonstrates the ability of language models to memorize and recall training data.

% \cite{zhang2024raft} shows a training strategy designed to enhance the model’s performance in answering questions.
% They also show that grounding the response to a piece of evidence improves model's effectiveness, similarly to a \emph{Proof}. 

% \cite{wu2024clasheval} highlights that there exists a tug-of-war between the strength of the model’s prior and the rate at which the model adheres to the retrieved documents.

% \cite{kandpal2023struggle} regards parametric knowledge.

\section{Conclusions}

In this paper, we aimed to systematically investigate the differences between LLM's base and instruct versions when used in RAG systems. 
Our findings reveal an unexpected outcome: base models exhibit superior performance on RAG tasks compared to their instructed and aligned counterparts. 
Further analysis indicates a tradeoff between accuracy and trustworthiness.
This tradeoff calls for novel evaluation methodologies for RAG pipelines and suggests the necessity for mechanisms that afford users greater control in managing this tradeoff in a more direct and explicit manner.

% Depending on the application the base model can be better than the instruct. Instruct models are useful without a template in RAG application when you need a model that you want to adhere to a specific instructions. Base models are better in extracting info. 

% Remember to talk about the cases without NO-RES in the instruction

% Instruct + Template are penalized because they usually prefer answering with long responses, which in most of cases are truncated due to the max_length on the generation.

\clearpage 
\newpage

\section{Limitations}
\label{sec:limitations}

Our study is subject to several limitations. 
Primarily, due to computational resource constraints, we did not evaluate LLMs with more than 8B parameters, which might offer additional insights into the effectiveness of base versus instruct versions for RAG applications. 
Similarly, we quantize models to 4-bit.
Additionally, our analysis could benefit from incorporating a broader range of datasets, particularly those that do not rely on Wikipedia as the primary knowledge source, or long-form QA datasets.

A critical limitation lies in the evaluation methodology. As noted in several studies \cite{katranidis2024faaf, yu2024evaluation}, verifying the presence of the ground truth answer within the generated response can sometimes inaccurately penalize correct answers. This typically occurs if the ground truth is not fully captured in the response, even after normalization. 

% However, since our datasets primarily require concise answers (e.g., the NQ-open dataset limits responses to five tokens) that are directly extractable from the documents, accuracy is generally considered an appropriate metric.

% Moreover, TriviaQA presents many different potential answers for each query, reaching in some cases up to 400 possible answers. This diversity helps reduce the number of times a generated response is considered incorrect even though it is semantically correct. This is evident in the results (Section \ref{sec:results}), where the accuracy on TriviaQA is greater in absolute terms compared to the NQ scores.

% \florin{About Template}
% This problem is especially notable in the NQ dataset where the LLM’s generation stops at 15 tokens. However, task instructions for NQ do specify that answers should be concise, ideally extracting a maximum of 5 tokens from the context (Figure ??).

% In addition to this, instruct models with templates underperform compared to their non-template counterparts, also in the TriviaQA dataset, where the maximum token generation limit is 50 (Table \ref{tab:trivia_first}). Despite the task instructions specifying brief or simple extraction responses, models with templates tend to generate longer answers. This tendency may be linked to their fine-tuning and alignment for conversational purposes, where verbosity can be advantageous to assist users. However, when templates are removed, these models still manage to answer accurately, often outperforming their templated versions.

\clearpage
\newpage

% \section*{Acknowledgments}

% Acks

\bibliography{biblio}

\clearpage 
\newpage

\appendix

\section{More Details on Datasets and Models}
\label{sec:appendix}

\subsection{Dataset}

We employ the NQ-open and TriviaQA-unfiltered datasets for our evaluations. For NQ-open we adopt the processing procedure of \citet{cuconasu2024power}, resulting in 2,889 test examples. For TriviaQA-unfiltered, we adhere to the validation and test split adopted in previous studies \cite{min-etal-2019-discrete, asai2023selfrag}, using 11,313 test queries for evaluation. The two datasets use the English Wikipedia dated 20 December 2018 as a knowledge source. Following the Dense Passage Retrieval (DPR) approach \cite{karpukhin-etal-2020-dense}, we split each Wikipedia article into non-overlapping passages of 100 words.

Both datasets feature questions that allow for multiple valid answers. These can range from synonymous terms, such as ``New York'' and ``NY'', to questions accepting multiple distinct correct answers. TriviaQA is notable for its diversity in acceptable responses, with some queries accommodating up to 400 valid answers. This variance significantly lowers the likelihood of a correct response being marked as incorrect purely based on phrasing variances, contributing to notably higher accuracy scores on TriviaQA than on NQ-open, as discussed in Section \ref{sec:results}.

 % In particular, TriviaQA presents many different potential answers for each query, reaching in some cases up to 400 possible answers. This diversity helps reduce the number of times a generated response is considered incorrect even though it is semantically correct. This is evident in the results (Section \ref{sec:results}), where the accuracy on TriviaQA is greater in absolute terms compared to the NQ scores.

\subsection{Retriever}
\label{sec:retriever_appendix}

The retriever used to select the top-$k$ documents is Contriever \cite{izacard2021contriever}, which is a BERT-based dense model trained using unsupervised contrastive loss. The embedding of each document and query is obtained by averaging the hidden
state of the last layer of the model.
For document retrieval from the corpus, we employ a FAISS index \cite{douze2024faiss, johnson2019billion} by using an inner product similarity metric (IndexFlatIP) with an exhaustive search.

To assess the availability of an answer within the provided documents, we compute the top-$k$ accuracy of the retriever. This metric evaluates how often the ground truth answer appears within the top-$k$ documents retrieved for a query. Scores can be seen in Table \ref{tab:contriever_acc}.

% \usepackage{booktabs}
\begin{table}[ht]
\centering
\caption{\textbf{Retriever top-$k$ accuracy.}}
\label{tab:contriever_acc}
\resizebox{\columnwidth}{!}{
\begin{tabular}{@{}c|ccccccc@{}}
\toprule
\textbf{Contriever} & \multicolumn{7}{c}{\textbf{\# Retrieved Documents}}   \\ \midrule
\textbf{Dataset}    & 1     & 2     & 3     & 4     & 5     & 8     & 10    \\ \midrule
NQ                  & 25.02 & 35.69 & 42.89 & 47.84 & 51.33 & 57.84 & 60.85 \\
TriviaQA            & 39.15 & 50.70 & 56.45 & 60.32 & 63.03 & 68.35 & 70.49 \\ \bottomrule
\end{tabular}
}
\end{table}

\begin{table}[ht]
\centering
\caption{\textbf{Closed-book Accuracy} of the models. The task instruction used in this case for both datasets can be seen in Figure \ref{fig:prompt_closed_book}.}
\label{tab:closed_book}
\resizebox{\columnwidth}{!}{
\begin{tabular}{@{}lc|c@{}}
\toprule
\multicolumn{1}{c|}{\textbf{Model}} &
  \textbf{\begin{tabular}[c]{@{}c@{}}NQ\\ Closed-Book\end{tabular}} &
  \textbf{\begin{tabular}[c]{@{}c@{}}TriviaQA\\ Closed-Book\end{tabular}} \\ \midrule
\multicolumn{1}{l|}{Llama 2 7B}       & \multicolumn{1}{c|}{\textbf{25.20}} & \textbf{55.57} \\
\multicolumn{1}{l|}{Llama 2 7B-C}     & \multicolumn{1}{c|}{16.51}          & 37.33          \\
\multicolumn{1}{l|}{Llama 2 7B-C + T} & \multicolumn{1}{c|}{13.12}          & 41.73          \\ \midrule
\multicolumn{1}{l|}{Llama 3 8B}       & \multicolumn{1}{c|}{22.01}          & \textbf{57.37} \\
\multicolumn{1}{l|}{Llama 3 8B-I}     & \multicolumn{1}{c|}{27.17}          & 25.46          \\
\multicolumn{1}{l|}{Llama 3 8B-I + T} & \multicolumn{1}{c|}{\textbf{27.80}} & 54.87          \\ \midrule
\multicolumn{1}{l|}{Mistral 7B}       & \multicolumn{1}{c|}{\textbf{28.11}} & \textbf{60.68} \\
\multicolumn{1}{l|}{Mistral 7B-I}     & \multicolumn{1}{c|}{18.76}          & 47.18          \\
\multicolumn{1}{l|}{Mistral 7B-I + T} & \multicolumn{1}{c|}{15.96}          & 49.30          \\ \midrule
\multicolumn{1}{l|}{Falcon 7B}        & \multicolumn{1}{c|}{\textbf{15.26}} & \textbf{41.12} \\
\multicolumn{1}{l|}{Falcon 7B-I}      & \multicolumn{1}{c|}{12.74}          & 27.83          \\ \bottomrule
\end{tabular}
}
\end{table}

% \usepackage{booktabs}
\begin{table*}[ht]
\centering
\caption{\textbf{Task Instruction I with No Rejection Percentage Accuracy} on NQ and TriviQA. In the No Rejection setting, we do not specify to answer with \emph{NO-RES} when the answer is not contained in the retrieved documents. An example of the task instruction adopted in these experiments can be seen in Figure \ref{fig:base_vs_instruct_no_rej}.}
\label{tab:nq_trivia_no_rejection}
\resizebox{2.09\columnwidth}{!}{
\begin{tabular}{@{}l|rrrrrrr|rrrrrrr@{}}
\toprule
 &
  \multicolumn{7}{c|}{\textbf{\begin{tabular}[c]{@{}c@{}}NQ\\ \# Retrieved Documents\end{tabular}}} &
  \multicolumn{7}{c}{\textbf{\begin{tabular}[c]{@{}c@{}}TriviaQA\\ \# Retrieved Documents\end{tabular}}} \\ \midrule
\multicolumn{1}{c|}{\textbf{Model}} &
  \multicolumn{1}{c}{1} &
  \multicolumn{1}{c}{2} &
  \multicolumn{1}{c}{3} &
  \multicolumn{1}{c}{4} &
  \multicolumn{1}{c}{5} &
  \multicolumn{1}{c}{8} &
  \multicolumn{1}{c|}{10} &
  \multicolumn{1}{c}{1} &
  \multicolumn{1}{c}{2} &
  \multicolumn{1}{c}{3} &
  \multicolumn{1}{c}{4} &
  \multicolumn{1}{c}{5} &
  \multicolumn{1}{c}{8} &
  \multicolumn{1}{c}{10} \\ \midrule
Llama 2 7B &
  \textbf{26.41} &
  \textbf{28.11} &
  \textbf{30.49} &
  \textbf{31.15} &
  \textbf{31.57} &
  \textbf{32.02} &
  \textbf{31.81} &
  \textbf{58.09} &
  \textbf{59.98} &
  \textbf{61.07} &
  \textbf{61.85} &
  \textbf{62.22} &
  \textbf{63.44} &
  \textbf{64.26} \\
Llama 2 7B-C &
  19.76 &
  25.20 &
  26.00 &
  27.62 &
  27.35 &
  27.76 &
  28.31 &
  47.29 &
  50.61 &
  52.93 &
  54.27 &
  55.06 &
  57.11 &
  57.71 \\
Llama 2 7B-C + T &
  4.05 &
  2.53 &
  1.56 &
  1.25 &
  1.11 &
  1.90 &
  3.63 &
  33.46 &
  39.70 &
  41.32 &
  41.34 &
  41.09 &
  39.01 &
  43.11 \\ \midrule
Llama 3 8B &
  \textbf{28.14} &
  \textbf{30.32} &
  \textbf{30.46} &
  \textbf{30.81} &
  28.66 &
  29.21 &
  28.94 &
  \textbf{49.65} &
  \textbf{57.45} &
  \textbf{61.07} &
  \textbf{63.17} &
  \textbf{64.72} &
  \textbf{66.53} &
  \textbf{66.80} \\
Llama 3 8B-I &
  13.36 &
  19.21 &
  23.19 &
  25.75 &
  \textbf{29.53} &
  \textbf{31.39} &
  \textbf{31.15} &
  4.05 &
  3.94 &
  5.23 &
  7.01 &
  18.41 &
  53.94 &
  61.46 \\
Llama 3 8B-I + T &
  14.88 &
  13.33 &
  12.81 &
  15.40 &
  20.32 &
  21.98 &
  21.98 &
  26.85 &
  37.59 &
  42.25 &
  44.53 &
  46.36 &
  53.15 &
  56.00 \\ \midrule
Mistral 7B &
  \textbf{25.48} &
  \textbf{26.48} &
  26.13 &
  26.83 &
  29.39 &
  29.15 &
  29.49 &
  \textbf{59.52} &
  \textbf{60.06} &
  \textbf{60.90} &
  \textbf{62.17} &
  \textbf{64.65} &
  \textbf{66.45} &
  \textbf{67.88} \\
Mistral 7B-I &
  21.53 &
  26.31 &
  \textbf{29.42} &
  \textbf{31.36} &
  \textbf{32.09} &
  \textbf{34.06} &
  \textbf{34.89} &
  51.87 &
  54.42 &
  56.19 &
  57.07 &
  58.07 &
  59.93 &
  61.25 \\
Mistral 7B-I+ T &
  19.56 &
  24.96 &
  26.76 &
  28.73 &
  27.83 &
  28.04 &
  36.55 &
  48.15 &
  51.99 &
  54.48 &
  55.36 &
  56.34 &
  58.86 &
  59.85 \\ \midrule
Falcon 7B &
  \textbf{16.58} &
  \textbf{19.45} &
  \textbf{19.80} &
  \textbf{21.46} &
  \textbf{21.08} &
  \textbf{21.08} &
  \textbf{21.26} &
  \textbf{41.66} &
  \textbf{43.55} &
  \textbf{43.40} &
  \textbf{43.92} &
  \textbf{45.04} &
  \textbf{45.62} &
  \textbf{45.54} \\
Falcon 7B-I &
  16.03 &
  18.28 &
  18.90 &
  19.38 &
  19.42 &
  19.38 &
  19.97 &
  33.99 &
  36.08 &
  36.78 &
  36.98 &
  37.89 &
  38.18 &
  38.43 \\ \bottomrule
\end{tabular}
}
\end{table*}

% \usepackage{booktabs}
\begin{table*}[ht!]
\centering
\caption{\textbf{Task Instruction II Answer Coherence Percentage Accuracy} on NQ and TriviaQA. It indicates the number of times the generated answer is contained in the generated \emph{Proof}, over the total number of instances.}
\label{tab:nq_trivia_proof_coherence}
\resizebox{2.09\columnwidth}{!}{
\begin{tabular}{@{}l|ccccccc|lllllll@{}}
\toprule
 &
  \multicolumn{7}{c|}{\textbf{\begin{tabular}[c]{@{}c@{}}NQ\\ \# Retrieved Documents\end{tabular}}} &
  \multicolumn{7}{c}{\textbf{\begin{tabular}[c]{@{}c@{}}TriviaQA\\ \# Retrieved Documents\end{tabular}}} \\ \midrule
\multicolumn{1}{c|}{\textbf{Model}} &
  1 &
  2 &
  3 &
  4 &
  5 &
  8 &
  10 &
  \multicolumn{1}{c}{1} &
  \multicolumn{1}{c}{2} &
  \multicolumn{1}{c}{3} &
  \multicolumn{1}{c}{4} &
  \multicolumn{1}{c}{5} &
  \multicolumn{1}{c}{8} &
  \multicolumn{1}{c}{10} \\ \midrule
Llama 2 7B &
  36.93 &
  41.16 &
  41.64 &
  41.43 &
  42.16 &
  \textbf{42.99} &
  \textbf{41.26} &
  37.22 &
  44.27 &
  44.68 &
  47.19 &
  48.00 &
  \textbf{50.35} &
  \textbf{47.14} \\
Llama 2 7B-C &
  \textbf{45.07} &
  \textbf{47.35} &
  \textbf{46.97} &
  \textbf{46.45} &
  \textbf{45.17} &
  40.53 &
  32.78 &
  \textbf{43.15} &
  \textbf{49.61} &
  \textbf{53.09} &
  \textbf{53.46} &
  \textbf{48.92} &
  40.14 &
  28.93 \\
Llama 2 7B-C + T &
  0 &
  0.59 &
  1.14 &
  0.07 &
  0.07 &
  0 &
  0 &
  0.05 &
  0.46 &
  0.75 &
  0.11 &
  0.04 &
  0 &
  0 \\ \midrule
Llama 3 8B &
  36.90 &
  41.36 &
  43.54 &
  \textbf{46.49} &
  46.80 &
  \textbf{51.78} &
  51.37 &
  \textbf{51.66} &
  \textbf{57.44} &
  \textbf{59.46} &
  \textbf{61.30} &
  \textbf{62.02} &
  \textbf{64.12} &
  \textbf{63.58} \\
Llama 3 8B-I &
  \textbf{42.02} &
  \textbf{51.92} &
  \textbf{50.71} &
  43.86 &
  \textbf{48.36} &
  48.29 &
  \textbf{53.10} &
  35.45 &
  50.55 &
  50.89 &
  48.91 &
  55.02 &
  55.28 &
  56.78 \\
Llama 3 8B-I + T &
  0.52 &
  0.03 &
  0.10 &
  0.14 &
  1.38 &
  0.28 &
  0.07 &
  1.10 &
  0.13 &
  0.12 &
  0.49 &
  1.59 &
  1.42 &
  0.5 \\ \midrule
Mistral 7B &
  36.73 &
  44.03 &
  46.56 &
  45.66 &
  51.47 &
  54.62 &
  53.31 &
  43.23 &
  52.16 &
  \textbf{56.07} &
  \textbf{58.48} &
  \textbf{60.82} &
  \textbf{66.66} &
  \textbf{68.04} \\
Mistral 7B-I &
  \textbf{50.12} &
  \textbf{53.24} &
  \textbf{54.38} &
  \textbf{55.49} &
  \textbf{57.39} &
  \textbf{57.74} &
  \textbf{56.07} &
  \textbf{51.08} &
  \textbf{53.52} &
  55.43 &
  57.53 &
  58.55 &
  59.68 &
  58.58 \\
Mistral 7B-I+ T &
  15.99 &
  10.90 &
  7.65 &
  5.75 &
  3.53 &
  4.12 &
  2.70 &
  16.82 &
  14.87 &
  11.42 &
  7.90 &
  5.70 &
  2.45 &
  2.16 \\ \midrule
Falcon 7B &
  1.87 &
  2.08 &
  3.32 &
  3.01 &
  3.81 &
  4.50 &
  1.15 &
  3.02 &
  3.70 &
  3.47 &
  4.52 &
  4.86 &
  3.17 &
  1.61 \\
Falcon 7B-I &
  \textbf{3.39} &
  \textbf{3.12} &
  \textbf{3.77} &
  \textbf{4.85} &
  \textbf{4.26} &
  \textbf{1.73} &
  \textbf{1.04} &
  \textbf{4.45} &
  \textbf{5.94} &
  \textbf{6.53} &
  \textbf{6.08} &
  \textbf{6.28} &
  \textbf{3.34} &
  \textbf{1.67} \\ \bottomrule
\end{tabular}
}
\end{table*}

\subsection{Generative Models}

We utilize publicly available, open-weight LLMs accessible via Hugging Face. All models are quantized to 4-bit using the bitsandbytes library\footnote{\href{https://huggingface.co/docs/bitsandbytes/main/en/index}{https://huggingface.co/docs/bitsandbytes/main/en/index}} to optimize computational efficiency. We perform all experiments with a single Nvidia RTX4090 GPU.

Here is a brief description of the main characteristics of each model:

\paragraph{Llama 2 7B.}
The 7B parameters versions\footnote{\href{https://huggingface.co/meta-llama/Llama-2-7b-hf}{https://huggingface.co/meta-llama/Llama-2-7b-hf}
\href{https://huggingface.co/meta-llama/Llama-2-7b-chat-hf}{https://huggingface.co/meta-llama/Llama-2-7b-chat-hf} } of the Llama 2 family \cite{touvron2023llama} are pre-trained on publicly available data and optimized for a range of natural language generation tasks. This series features a context length of 4096 tokens, and the 7B version employs multi-query attention (MQA) \cite{shazeer2019fast} to enhance processing efficiency and response quality.

\paragraph{Llama 3 8B.}
The Llama 3 series\footnote{\href{https://llama.meta.com/llama3/}{Llama 3 Blog Page}} builds on the architecture and improvements of its predecessors, offering models with 8B parameters\footnote{\href{https://huggingface.co/meta-llama/Meta-Llama-3-8B}{https://huggingface.co/meta-llama/Meta-Llama-3-8B}
\href{https://huggingface.co/meta-llama/Meta-Llama-3-8B-Instruct}{https://huggingface.co/meta-llama/Meta-Llama-3-8B-Instruct}}. It employs group-query attention (GQA) \cite{ainslie2023gqa} and extends the context length to 8192 tokens, thus facilitating enhanced language generation across a broad range of tasks.

\paragraph{Mistral 7B.}
Developed as a highly efficient model with 7B parameters\footnote{\href{https://huggingface.co/mistralai/Mistral-7B-v0.1}{hhttps://huggingface.co/mistralai/Mistral-7B-v0.1}
\href{https://huggingface.co/mistralai/Mistral-7B-Instruct-v0.1}{https://huggingface.co/mistralai/Mistral-7B-Instruct-v0.1}
}, Mistral \cite{jiang2023mistral} focuses on delivering high performance and accuracy in text generation. It uses GQA and sliding window attention with an 8192-token context length.

\paragraph{Falcon 7B.}
The Falcon 7B\footnote{\href{https://huggingface.co/tiiuae/falcon-7b}{https://huggingface.co/tiiuae/falcon-7b}
\href{https://huggingface.co/tiiuae/falcon-7b-instruct}{https://huggingface.co/tiiuae/falcon-7b-instruct}
} is the smallest model of the Falcon series \cite{almazrouei2023falcon} and
was trained on the RefinedWeb dataset \cite{penedo2023refined}—a large, filtered,
and deduplicated corpus. Similarly to Llama2 7B, it uses MQA, but with a smaller context length of 2048 tokens. Unlike the other models, the instruct version of Falcon 7B was not specifically trained using a fixed template, which is why no separate ``instruct + template'' variant is listed in any figure or table.
\\ \\
The closed-book accuracy of the models is detailed in Table \ref{tab:closed_book}. In this scenario, models are evaluated without any documents in their prompt, necessitating a modification to Task Instruction I. An example of this modified task instruction can be viewed in Figure \ref{fig:prompt_closed_book}.

\section{Further Analysis with Task Instruction II}

In this section, we examine more in detail whether models can justify their answers with \emph{Proof}. We specifically investigate whether even the base models can adhere to instructions and provide accurate \emph{Proof} for their responses.

Table \ref{tab:nq_trivia_proof_coherence} shows the percentage of instances where the generated answer, whether correct or incorrect, is included in the generated \emph{Proof}, which we will refer to as ``coherence''. The inclusion of the answer in the \emph{Proof} indicates that the model's responses align with the information provided in the documents. This measure, however, does not necessarily reflect answer correctness, as it only assesses coherence with the documented evidence.

As observed in Table \ref{tab:nq_trivia_proof_coherence}, base models are ``coherent'' with their answers, often outperforming their instruct counterparts. Mistral notably achieves the highest coherence score, reaching 68\% with 10 retrieved documents, while Falcon exhibits the lowest, often failing to provide any \emph{Proof} at all. Even the instruct version of Falcon typically offers only the direct answer without supporting evidence.

However, as discussed in the limitations section (Section \ref{sec:limitations}), the presence of an answer in the \emph{Proof} does not guarantee its ``coherence'' accuracy. An answer may be included, but the \emph{Proof} might not actually derive from the provided documents. Additionally, even if the answer is present in the \emph{Proof}, it may not be recognized as valid due to the inclusion of additional text in the response. For example, if a model begins its response with ``The answer to the question is...'' and then reports an answer that is technically part of the \emph{Proof}, this response might still be deemed invalid because the introductory phrase does not originate from the context documents.

% However, as discussed in the limitations section (Section \ref{sec:limitations}), the presence of an answer in the \emph{Proof} does not guarantee its ``coherence'' accuracy. For example, the answer may be included but the \emph{Proof} is not extracted from one of the provided documents.
% Or the answer, although present in the \emph{Proof}, may not be recognized as such due to the inclusion of additional text in the response, which is not present in the documents. For instance, if a model starts its response with "The answer to the question is..." and subsequently provides an answer that is included in the \emph{Proof}, this answer might be considered invalid because the introductory phrase is not part of the \emph{Proof}.

% an answer might appear in the \emph{Proof}, but the extractive \emph{Proof} may not conclusively support the correctness of the response.

\section{Further Analysis on the Negative Rejection Rates}

\begin{figure*}[ht]
    \centering
    \includegraphics[width=2.09\columnwidth]{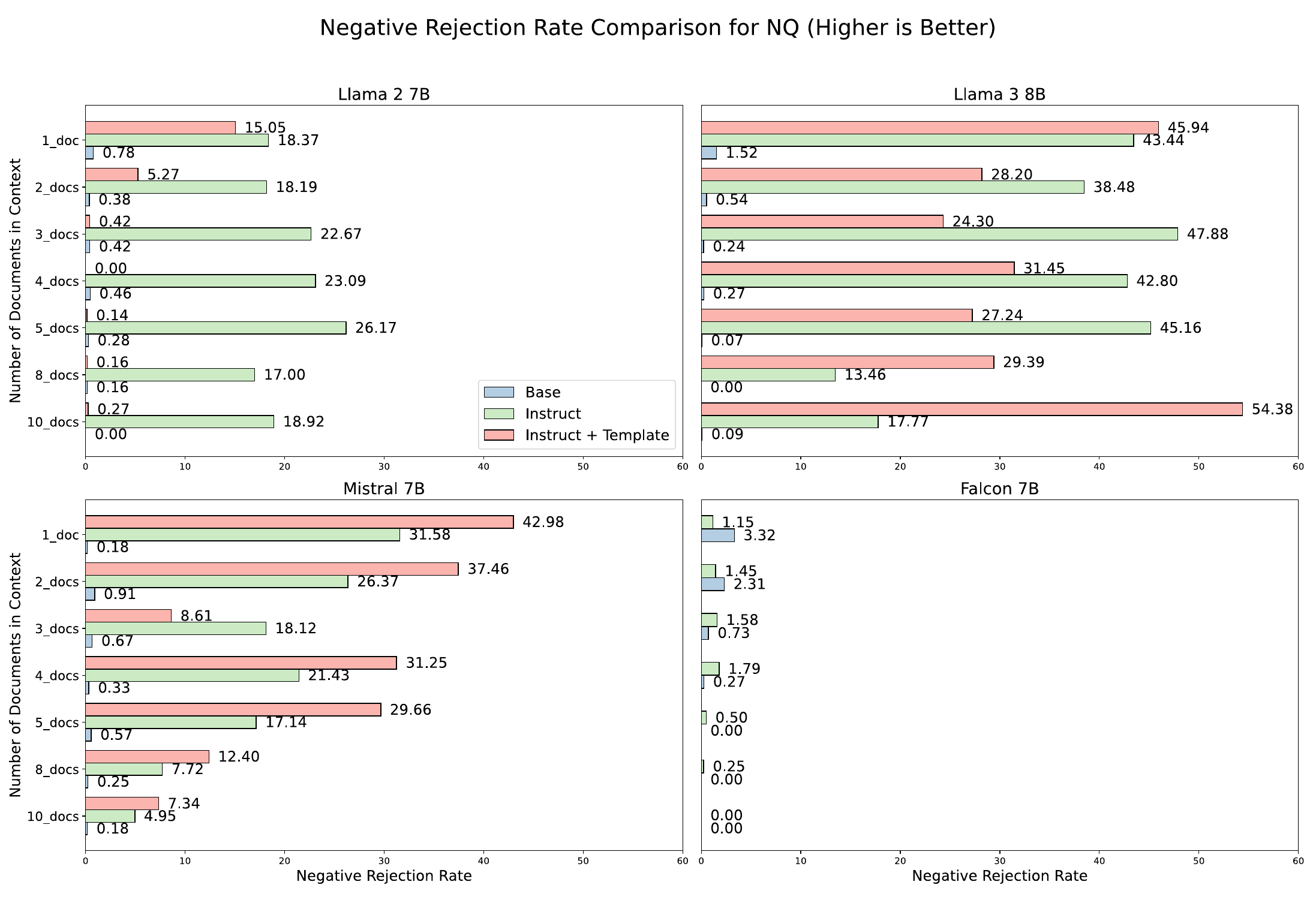}
  \caption{\textbf{Negative Rejection Comparison for NQ.} Reported is the negative rejection rate, that is, the number of times the model answers \emph{NO-RES} when the correct answer is not in the context, divided by the number of times the answer is indeed missing.
  Instruct models are much more effective at detecting such cases and following the instructions provided.}
  \label{fig:all_neg_rej_NQ}
\end{figure*}

\begin{figure*}[ht]
    \centering
    \includegraphics[width=2.09\columnwidth]{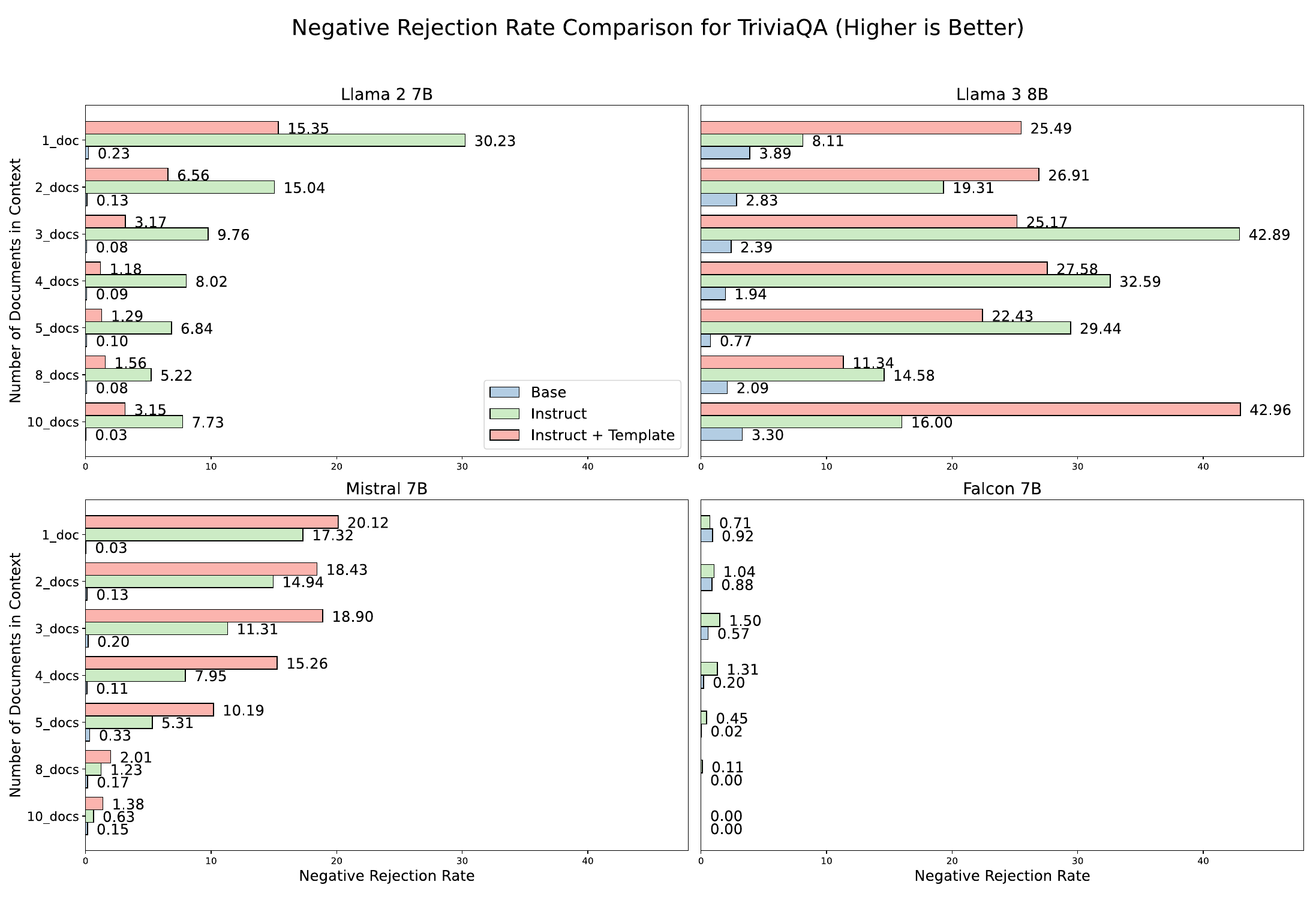}
  \caption{\textbf{Negative Rejection Comparison for TriviaQA.} Reported is the negative rejection rate, that is, the number of times the model answers \emph{NO-RES} when the correct answer is not in the context, divided by the number of times the answer is indeed missing.
  Instruct models are much more effective at detecting such cases and following the instructions provided.}
  \label{fig:all_neg_rej_TriviaQA}
\end{figure*}

In Section \ref{sec:neg_rej}, we discussed the negative rejection ability of Llama 2 on TriviaQA. This section extends the discussion to the behavior of other models on both datasets, NQ and TriviaQA. From Tables \ref{fig:all_neg_rej_NQ} and \ref{fig:all_neg_rej_TriviaQA}, it is evident that the instruct models generally exhibit higher negative rejection rates compared to their base counterparts. However, each model follows its own trend.

Llama 2 and Mistral demonstrate similar behaviors in that, as the number of documents in the context increases, their tendency to answer with \emph{NO-RES} decreases. Notably, the instruct version with a template for Llama 2 rarely rejects to respond when the answer is not present in the context documents. In contrast, the Mistral instruct version with a template shows a significantly stronger negative rejection ability than without the template, reaching up to 42.98\% score with one retrieved document on NQ.

Llama 3 exhibits a distinct trend, maintaining a mean negative rejection rate of 34\% with the template, and 35.57\% without. However, when the model is not using the template, similar to Llama 2 and Mistral, the rejection rates for Llama 3 decline with an increase in document count. Indeed, it shows a significant reduction of 31.7 (-70\%) passing from 45.16\% to 13.46\% when the number of retrieved documents increases from 5 to 8 on NQ; while for TriviaQA the rate drops by 14.86 (-50\%) in the same situation. 

Falcon models show the least tendency to respond with \emph{NO-RES}, particularly in the instruct and instruct-template settings, where rejection rates are consistently low or even non-existent in some configurations. This behavior indicates a propensity to generate answers even when the information is not present, potentially leading to higher rates of hallucination. The primary reason for this is Falcon's difficulty in following instructions effectively.

\begin{figure*}[htbp]
    % \centering
    \begin{tabular}
    {|p{0.47\textwidth}|p{0.47\textwidth}|}
        \hline
        \centering
        \textbf{Mistral 7B} \arraybackslash  & \centering \textbf{Mistral 7B-Instruct + Template} \arraybackslash \\
        \hline
        {You are given a question and you MUST respond by EXTRACTING the answer (max 5 tokens) from one of the provided documents. If none of the documents contain the answer, respond with NO-RES.
        \newline
        \textbf{START example}\newline
        \textbf{Document} [209707](Title: Ancient Egyptian technology) Evidence indicates that Egyptians made use of potter's wheels in the manufacturing of pottery from as early as the 4th Dynasty. Chariots, however, are only believed to have been introduced by the invasion of the Hyksos in the Second Intermediate period; during the New Kingdom era, chariotry became central to Egypt's military.\newline
        \textbf{Question:} when was the potter's wheel first used in egypt\newline
        \textbf{Answer:} 4th Dynasty\newline
        \textbf{Proof:} Evidence indicates that Egyptians made use of potter's wheels in the manufacturing of pottery from as early as the 4th Dynasty.\newline
        \textbf{END example}
        } 
        & \emph{[INST]} You are given a question and you MUST respond by EXTRACTING the answer (max 5 tokens) from one of the provided documents. If none of the documents contain the answer, respond with NO-RES.
        \newline
        \textbf{START example}\newline
        \textbf{Document} [209707](Title: Ancient Egyptian technology) Evidence indicates that Egyptians made use of potter's wheels in the manufacturing of pottery from as early as the 4th Dynasty. Chariots, however, are only believed to have been introduced by the invasion of the Hyksos in the Second Intermediate period; during the New Kingdom era, chariotry became central to Egypt's military.\newline
        \textbf{Question:} when was the potter's wheel first used in egypt\newline
        \textbf{Answer:} 4th Dynasty\newline
        \textbf{Proof:} Evidence indicates that Egyptians made use of potter's wheels in the manufacturing of pottery from as early as the 4th Dynasty.\newline
        \textbf{END example} \\
        \hline
        \multicolumn{2}{|p{0.94\textwidth}|}{
            \textbf{Documents:}
            
            \textbf{Document [12919694]}(Title: User State Migration Tool) User State Migration Tool The User State Migration Tool (USMT) is a Microsoft command line utility program intended to allow advanced users, comfortable with Scripting languages, to transfer files and settings between PCs. This task is also performed by \textcolor{correct}{Windows Easy Transfer}, recommended for general users...
            
            \textbf{Document [12919696]}(Title: User State Migration Tool) Settings Transfer Wizard and Windows Easy Transfer. USMT is designed to perform mass migrations and allows customizations designed for corporate environments...
            
            \textbf{Document [12919695]}(Title: User State Migration Tool) there have been several attempts to provide access to its useful functionality by creating GUI wrappers for it...\newline

            \textbf{Question:} what tool is available from microsoft as an alternative to the usmt for smaller migrations
        } \\
        \hline
        \textbf{Answer:} \textcolor{correct}{Windows Easy Transfer}
        \newline
        \textbf{Proof:} User State Migration Tool The User State Migration Tool (USMT) is a Microsoft command line utility program intended to allow advanced users, comfortable with Scripting languages, to transfer files and settings between PCs. This task is also performed by \textcolor{correct}{Windows Easy Transfer}, recommended for general users...
        & 
        
        \textbf{Answer:} \emph{[/INST]} \textcolor{red}{NO-RES.}
        \\
        \hline
    \end{tabular}
    \caption{\textbf{Base vs. Instruct + Template under Task Instruction II on NQ}. This comparison of responses between the base and instruct + template versions of Mistral 7B illustrates an example where the base model correctly identifies the answer, while the instruct + template version erroneously opts for a \emph{NO-RES} response, despite the correct answer being present in the documents. \emph{Italic} text denotes the template.}
    \label{fig:base_vs_instruct_proof_mistral_nq}
\end{figure*}

\begin{figure*}[htbp]
    % \centering
    \begin{tabular}
    {|p{0.47\textwidth}|p{0.47\textwidth}|}
        \hline
        \centering
        \textbf{Llama 2 7B} \arraybackslash  & \centering \textbf{Llama 2 7B-Chat + Template} \arraybackslash \\
        \hline
        {\emph{} \newline You are given a question and you MUST respond by EXTRACTING the answer from one of the provided documents. If none of the documents contain the answer, respond with NO-RES.
        \newline
        \textbf{START example}\newline
        \textbf{Document} [209707](Title: Ancient Egyptian technology) Evidence indicates that Egyptians made use of potter's wheels in the manufacturing of pottery from as early as the 4th Dynasty. Chariots, however, are only believed to have been introduced by the invasion of the Hyksos in the Second Intermediate period; during the New Kingdom era, chariotry became central to Egypt's military.\newline
        \textbf{Question:} when was the potter's wheel first used in egypt\newline
        \textbf{Answer:} 4th Dynasty\newline
        \textbf{Proof:} Evidence indicates that Egyptians made use of potter's wheels in the manufacturing of pottery from as early as the 4th Dynasty.\newline
        \textbf{END example}
        } 
        & \emph{[INST] <<SYS>>} \newline
        You are given a question and you MUST respond by EXTRACTING the answer from one of the provided documents. If none of the documents contain the answer, respond with NO-RES.
        \newline
        \textbf{START example}\newline
        \textbf{Document} [209707](Title: Ancient Egyptian technology) Evidence indicates that Egyptians made use of potter's wheels in the manufacturing of pottery from as early as the 4th Dynasty. Chariots, however, are only believed to have been introduced by the invasion of the Hyksos in the Second Intermediate period; during the New Kingdom era, chariotry became central to Egypt's military.\newline
        \textbf{Question:} when was the potter's wheel first used in egypt\newline
        \textbf{Answer:} 4th Dynasty\newline
        \textbf{Proof:} Evidence indicates that Egyptians made use of potter's wheels in the manufacturing of pottery from as early as the 4th Dynasty.\newline
        \textbf{END example}\newline
        \emph{<</SYS>>} \\
        \hline
        \multicolumn{2}{|p{0.94\textwidth}|}{
            \textbf{Documents:}
            
            \textbf{Document [1]}(Title: Batman Returns) the Penguin. We didn't really officially cast it, but for a short nasty little guy, it's a short list. I ended up writing the character for Danny DeVito. \textcolor{correct}{Burgess Meredith} (who portrayed the Penguin in the 1960s TV series "Batman") was cast for a little cameo as Tucker Cobblepot...
            
            \textbf{Document [2]}(Title: Batman: Mystery of the Batwoman) This is the only time in the DC animated universe that Paul Williams did not voice the Penguin...
            
            \textbf{Document [3]}(Title: The Penguin's a Jinx) The Penguin goes to Wayne Manor and returns the actress. He then uses his gas-umbrella to knock out anyone inside the statues...\newline

            \textbf{Question:} Who played the part of `The Penguin' in the TV series `Batman'?
        } \\
        \hline
        \textbf{Answer:} \textcolor{correct}{Burgess Meredith} 
        \newline
        \textbf{Proof:} Document [1](Title: Batman Returns) the Penguin. We didn't really officially cast it, but for a short nasty little guy, it's a short list. I ended up writing the character for Danny DeVito. \textcolor{correct}{Burgess Meredith} (who portrayed the Penguin in the 1960s TV series "Batman") was cast for a little cameo as Tucker Cobblepot...
        & 
        
        \textbf{Answer:} \emph{[/INST]} \textcolor{red}{Danny DeVito}
        \newline
        \textbf{Proof:} Document [1](Title: Batman Returns) the Penguin. We didn't really officially cast it, but for a short nasty little guy, it's a short list. I ended up writing the character for \textcolor{red}{Danny DeVito}.
        \\
        \hline
    \end{tabular}
    \caption{\textbf{Base vs. Instruct + Template under Task Instruction II on TriviaQA}. This comparison of responses between the base and instruct + template versions of Llama 2 7B illustrates an example where the base model correctly identifies the answer, while the instruct + template version inaccurately attributes the answer to a different actor. Nevertheless, in both cases, the answers are ``coherent'' with the \emph{Proof} since each evidence contains the generated answer. \emph{Italic} text denotes the template.}
    \label{fig:base_vs_instruct_proof}
\end{figure*}

\begin{figure*}[htbp]
    % \centering
    \begin{tabular}{|p{0.47\textwidth}|p{0.47\textwidth}|}
        \hline
        \centering
        \textbf{Falcon 7B} \arraybackslash  & \centering \textbf{Falcon 7B-Instruct} \arraybackslash \\
        \hline
        You are given a question and you MUST respond by EXTRACTING the answer (max 5 tokens) from one of the provided documents. & You are given a question and you MUST respond by EXTRACTING the answer (max 5 tokens) from one of the provided documents. \\
        \hline
        \multicolumn{2}{|p{0.94\textwidth}|}{
            \textbf{Documents:}
            
            \textbf{Document [1590238]}(Title: 1970s in film) Thanks to a stellar cast, experienced director, and a poignant story, "Network" became one of the largest critical successes of 1976. Another film, \textcolor{correct}{Rocky}, about a clubhouse boxer (played by Sylvester Stallone) who is granted a world championship title fight won the Best Picture Academy Award that year. The film also became a major commercial success and spawned four sequels through the rest of the 1970s and 1980s...
            
            \textbf{Document [2832642]}(Title: Ned Beatty) 1975. Beatty received his first Academy Award nomination for Best Supporting Actor category for the acclaimed film "Network" (1976), portraying a television network's bombastic but shrewd chairman of the board who convinces the mad Howard Beale character (portrayed by Peter Finch) that corporation-led global dehumanization is not only inevitable, but is also a good thing...\newline

            \textbf{Question:} who won the oscar for best picture in 1976
        } \\
        \hline
        \textbf{Answer:} \textcolor{correct}{Rocky} & 
        
        \textbf{Answer:} \textcolor{red}{"Network"} \\
        \hline
    \end{tabular}
    \caption{\textbf{Base vs. Instruct under Task Instruction I with No Rejection on NQ}. This figure presents responses under a No Rejection setting, where models are not tasked with responding with \emph{NO-RES} if the answer is not contained in the retrieved documents.
    It compares the base and instruct versions of Falcon 7B. In this instance, the base model accurately identifies ``Rocky'' as the Oscar winner for Best Picture in 1976, while the instruct version incorrectly cites "Network".}
    \label{fig:base_vs_instruct_no_rej}
\end{figure*}

\begin{figure*}[htbp]
    % \centering
    \begin{tabular}{|p{0.47\textwidth}|p{0.47\textwidth}|}
        \hline
        \centering
        \textbf{Llama 3 8B} \arraybackslash  & \centering \textbf{Llama 3 8B-Instruct + Template} \arraybackslash \\
        \hline
        {\emph{}\newline \newline You are given a question and you MUST respond with a short answer based on your internal knowledge. If you do not know the answer, please respond with NO-RES.} & \emph{<|start\_header\_id|>system<|end\_header\_id|>} \newline \newline
        You are given a question and you MUST respond with a short answer based on your internal knowledge. If you do not know the answer, please respond with NO-RES.\emph{<|eot\_id|>\newline<|start\_header\_id|>user<|end\_header\_id|>} \\
        \hline
        \multicolumn{2}{|p{0.94\textwidth}|}{
            \textbf{Question:} In which US city did the 2004 remake of the film Alfie take place?
        } \\
        \hline
        \textbf{Answer:} \textcolor{correct}{New York City} & 
        
        \textbf{Answer:} \emph{<|eot\_id|><|start\_header\_id|>assistant\newline<|end\_header\_id|>} \textcolor{red}{London!} \\
        \hline
    \end{tabular}
    \caption{\textbf{Base vs. Instruct + Template under Closed-Book QA on TriviaQA}. This figure compares responses from the base and instruct + template versions of Llama 3 8B for a question in a closed-book setting, where no additional documents are provided. The example demonstrates how the base model accurately identifies ``New York City'' as the setting of the 2004 remake of the film Alfie, whereas the instruct + template version erroneously claims the location as ``London''. \emph{Italic} text denotes the template.}
    \label{fig:prompt_closed_book}
\end{figure*}

\end{document}